\documentclass{bmvc2k}
\pdfoutput=1

\title{Self-Supervised Real-time Video Stabilization}

\addauthor{Jinsoo Choi}{jinsc37@kaist.ac.kr}{1}
\addauthor{Jaesik Park}{jaesik.park@postech.ac.kr}{2}
\addauthor{In So Kweon}{iskweon77@kaist.ac.kr}{1}

\addinstitution{
 KAIST\\
 Republic of Korea
}
\addinstitution{
 POSTECH\\
 Republic of Korea
}

\runninghead{Choi, Park, Kweon}{Self-Supervised Real-time Video Stabilization}

\def\etal{\emph{et al}\bmvaOneDot}

\usepackage{times}
\usepackage{epsfig}
\usepackage{graphicx}
\usepackage{amsmath}
\usepackage{amssymb}
\usepackage{balance}
\usepackage{comment}
\usepackage[outercaption]{sidecap}

\begin{document}

\maketitle

\begin{abstract}
Videos are a popular media form, where online video streaming has recently gathered much popularity.
In this work, we propose a novel method of real-time video stabilization - transforming a shaky video to a stabilized video as if it were stabilized via gimbals in real-time.
Our framework is trainable in a self-supervised manner, which does not require data captured with special hardware setups (i.e., two cameras on a stereo rig or additional motion sensors).
Our framework consists of a transformation estimator between given frames for global stability adjustments, followed by scene parallax reduction module via spatially smoothed optical flow for further stability.
Then, a margin inpainting module fills in the missing margin regions created during stabilization to reduce the amount of post-cropping.
These sequential steps reduce distortion and margin cropping to a minimum while enhancing stability.
Hence, our approach outperforms state-of-the-art real-time video stabilization methods as well as offline methods that require camera trajectory optimization. 
Our method procedure takes approximately 24.3 ms yielding 41 fps regardless of resolution (e.g., 480p or 1080p).
\end{abstract}

\section{Introduction}
\begin{SCfigure}[][t]
    \centering
    \includegraphics[width=0.6\linewidth,keepaspectratio]{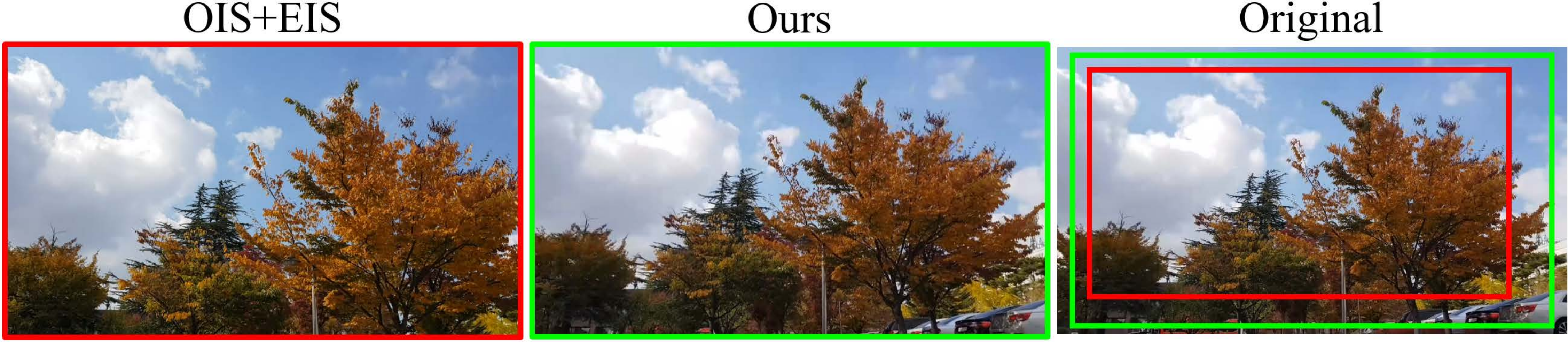}
    \caption{Example of the cropping ratios among OIS+EIS, our method and original frames taken by a Galaxy S10.
    OIS+EIS exhibits \emph{zoom-in effects}.
    }
    \vspace{-3mm}
    \label{ois}
\end{SCfigure}

Due to the recent popularity in social networking services (SNS), videos have become a popular media form, demanding higher visual quality as time progresses.
Recently, cameras (including smartphones) make use of hardware configurations to produce stabilized videos.
One such method is the Optical Image Stabilizer (OIS), which negates instability caused by hand movements via adjusting the optical lens positions. 
Another mechanism is the Electronic Image Stabilizer (EIS) that is designed to compensate for more substantial motion.
However, these methods require specialized motion-sensing hardware synchronized with image capture, and may lead to significant cropping of the frame boundaries of the original video, which results in an inevitable zoom-in effect. 
Fig.~\ref{ois} shows a real example of a video captured with OIS and EIS. 

\begin{SCfigure}[][t]
    \centering
    \includegraphics[width=0.6\linewidth,keepaspectratio]{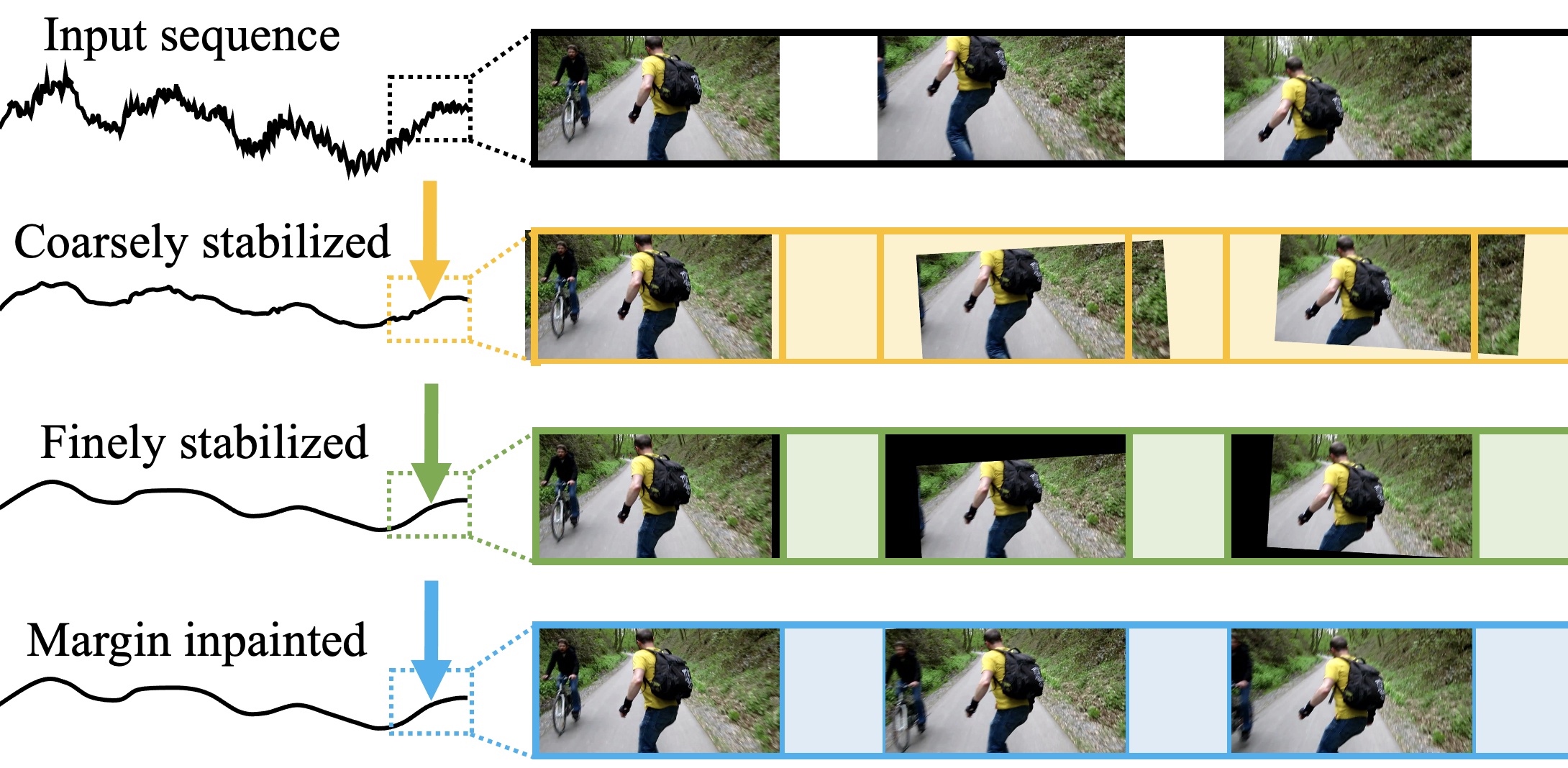}
    \caption{Illustration of approach. 
    Given a shaky camera path (black line profile), our method applies global transforms(top row).
    Then, spatially smoothed flow warping is applied for further stability(2nd row).
    Lastly, the image margins are inpainted to minimize missing regions (last row).
    }
    \vspace{-3mm}
    \label{teaser}
\end{SCfigure}

To cover such limitations, software approaches have been developed typically for offline purposes, namely post-processing of existing videos~\cite{liu2013bundled,liu2014steadyflow,grundmann2011auto}.
Offline approaches have shown success with stabilization quality, but the computational speed is an issue.
More recently, due to the growing popularity in online streaming, works on real-time methods~\cite{liu2017codingflow,liu2016meshflow,wang2018deep,xu2018deep} have been introduced. 
Especially, deep learning-based methods~\cite{wang2018deep,xu2018deep} utilize the fast feed-forward mechanism and a supervised approach to realize real-time performance. 
Compared to offline methods, however, the real-time approaches exhibit a slight loss of stability and increased distortion artifacts. 
Moreover, supervised learning methods require capturing simultaneous ground truth footages for supervised learning, which is a time-consuming process. 

A desirable video stabilization can be characterized by minimal cropping, low visual distortion, and high stability.
In this work, we propose a deep learning-based approach to enable real-time video stabilization with these aspects.
The idea is to introduce a cascade module which consists of a coarse (global) stabilizer and a fine (detail) stabilizer to handle complex camera motion in the wild effectively. 
The coarse stabilizer estimates the rigid inter-frame transformations among the multiple (unstable) frame inputs for global adjustment of subsequent frames. 
Next, a fine stabilizer applies the spatially smoothed optical flow between input frames to handle parallax and spatially varying instability.
Both stabilizers use moving average filters to suppress noisy camera motion without explicitly optimizing camera trajectories.
After the stabilization step, the inpainting module fills in the blank margins induced by the stabilization process. 
Fig.~\ref{teaser} shows an illustration of the process.

A notable aspect of our method is that our approach does not require any ground truth videos for deep architecture training, but yet manages to produce strong stabilization effects in a self-supervised manner. 
This distinguishes our method with supervised methods~\cite{wang2018deep,xu2018deep} that require sets of two videos, one of which is unstable while the other is physically stabilized with gimbals, captured simultaneously.
The proposed network is fully convolutional and is trained end-to-end, where our model achieves real-time regardless of input video size (e.g., 480p or 1080p) with a single GPU.
The contributions of our work are listed as follows.
\begin{itemize}
    \item We propose a real-time approach for video stabilization. 
    The input video is stabilized via a single feed-forward through the constituent modules. 
    \item We introduce a self-supervised training method for video stabilization. Therefore, existing videos can be used for network training.
    \item Our method induces low levels of cropping, low visual distortion, and high stability. The proposed approach outperforms the state-of-the-art methods both quantitatively and qualitatively.
\end{itemize}

\section{Related Work}
Video stabilization approaches consist of offline and real-time methods. 
Offline methods are characterized by strong stabilization effects while having a high computational load.
On the other hand, real-time methods are computationally efficient but convey weaker stabilization effects compared to offline methods.
We briefly address offline and real-time approaches in the following.

Previous works on offline methods deal with techniques including Structure from Motion (SfM)~\cite{liu2009content}, depth information~\cite{liu2012video}, 3D plane constraints~\cite{zhou2013plane}, 3D reconstruction~\cite{buehler2001non}, light fields~\cite{smith2009light}, gyroscopes~\cite{bell2014non,karpenko2011digital,ovren2015gyroscope}, and partial 3D information~\cite{goldstein2012video,liu2011subspace,liu2013joint}.
Image-based methods have also shown sufficient quality~\cite{chen2008capturing,gleicher2007re,huang2018encoding,matsushita2006full,yu2020learning}.
Grundmann~\etal~\cite{grundmann2011auto} apply L1-norm optimization for camera path computation and extend the approach to handle rolling shutter effects~\cite{grundmann2012calibration}.
Liu~\etal~\cite{liu2013bundled} model camera paths for each image patches, and extend the idea to model the entire pixel profiles~\cite{liu2014steadyflow}. 
This work shows that spatially smoothed flow is useful for stabilizing frames, which our approach builds upon. 
Recently, Choi and Kweon~\cite{choi2020deep} introduce an iterative frame interpolation approach to video stabilization.

Real-time methods typically employ more efficient computations while using historical frames to estimate stabilization parameters.
Liu~\etal~\cite{liu2016meshflow} proposes computing sparse mesh profiles via applying median filters to pixel profiles.
This idea is extended further to video coding~\cite{liu2017codingflow}.
Limitations of these methods are the slight wobbling artifacts and relatively low stabilization quality compared to offline methods.
Recently, deep learning-based algorithms~\cite{liu2021hybrid} take advantage of fast feed-forward capabilities.
Wang~\etal~\cite{wang2018deep} propose a supervised learning approach to video stabilization by defining the novel stability and temporal loss terms. 
Similarly, Xu~\etal~\cite{xu2018deep} proposes training an adversarial network in a supervised manner, which estimates transformation parameters to generate stabilized frames. Compared with those approaches, our approach does not require stable videos for training.
Recently, Yu and Ramamoorthi~\cite{yu2020learning} propose a learning method using principal components of optical flow information. Our approach estimate smooth warping field via neural network and produces high-quality stabilization.

The core module of our pipeline is related to deep image tranformation estimation and image inpainting. Deep homography estimation~\cite{detone2016deep,nguyen2018unsupervised} enables transformation matrix estimation between two frames without explicitly extracting image features. 
On the other hand, advances in image inpainting tasks employ specific deep architecture components for natural inpainting quality~\cite{liu2018image,yu2018free}.
Although image boundary inpainting is a difficult task, our deep margin inpainter utilizes multiple adjacent frames to fill in frame boundaries.
\begin{figure*}[ht]
    \includegraphics[width=1\linewidth,keepaspectratio]{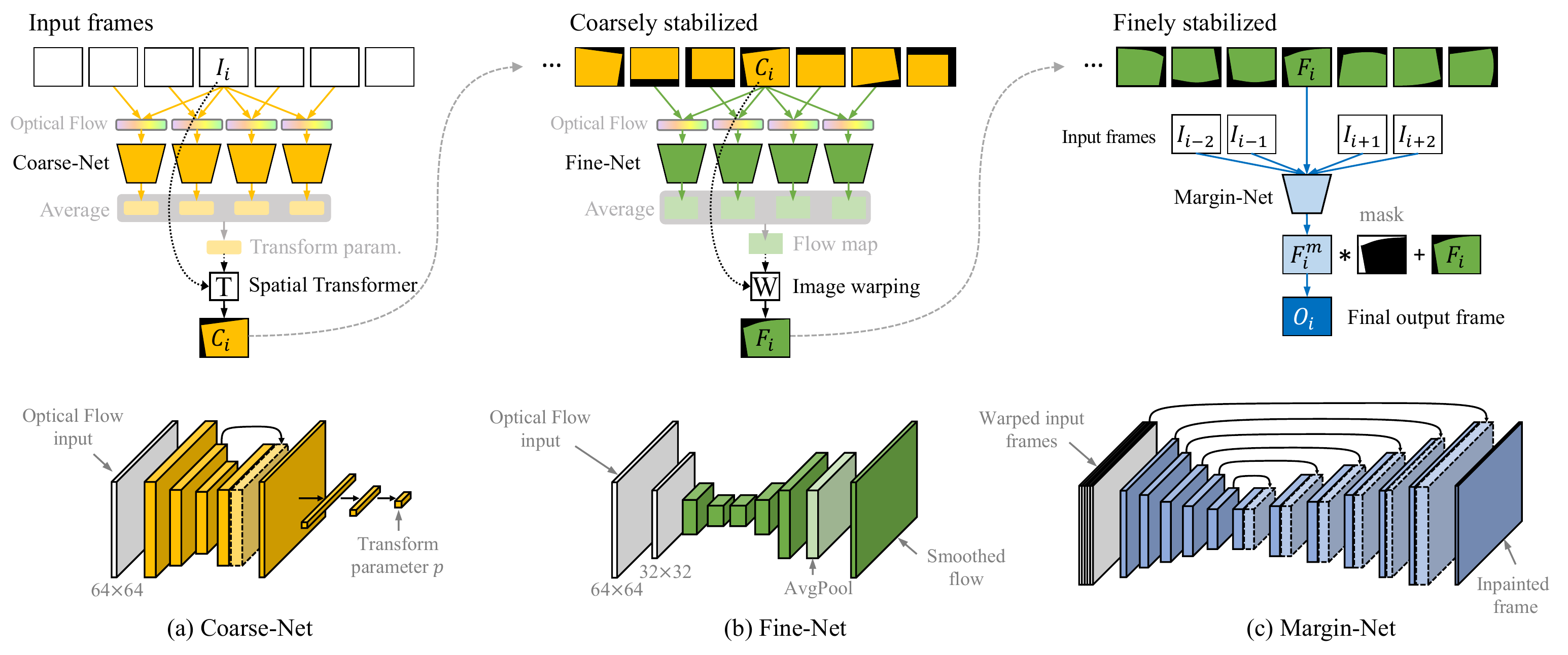}
    \caption{An overview of the Coarse-Net, Fine-Net and Margin-Net during testing.
    The Coarse-Net computes the transform parameters among input frames with respect to the center image $I_i$, which are then averaged to transform $I_i$ to $C_i$.
    The Fine-Net compute the mean of the flow maps to warp $C_i$, generating $F_i$.
    Finally, the Margin-Net takes $F_i$ and neighboring input frames to generate $F_i^m$, which is applied with the margin mask to generate the final output frame $O_i$.
    }
    \vspace{-3mm}
    \label{overview}
\end{figure*}

\section{Architecture}
We aim to achieve high-quality stabilization and high computational speed via three sequential modules, namely the coarse stabilizer, fine stabilizer, and margin inpainter. 
The overall process of our method is illustrated in Fig.~\ref{overview}.
The coarse (first) stabilizer estimates the rigid transformations among frames to globally transform a frame for stabilization.
The fine (second) stabilizer estimates the spatially smoothed flow among frames to adjust the remaining instability.
Finally, the margin inpainter fills in the frame boundaries to reduce the need for cropping.
Unlike prior work that runs offline, our approach does not involve any optimization, camera pose estimation.
Instead, sequential feed-forward passes of the coarse and fine networks produce the stabilized frames, which has been shown to be effective~\cite{yu2020learning}.



\vspace{-2mm}
\subsection{Coarse stabilizer}
\vspace{-2mm}
Our approach first estimates the image transformations between video frames.\footnote{We empirically found the 3-DoF rigid transformation $\bigl[ \begin{smallmatrix}R({\theta)} & \mathbf{t}\\ 0 & 1\end{smallmatrix}\bigr]$ is surprisingly effective for video stability, compared with transformations having higher DoF. The supplement summarizes our experiment results on other type of transformations that include scale and/or shear. Please note that  the combination of Coarse stabilizer and Fine stabilizer and their moving average can effectively handle challenging motions (zoom, parallax, and so on), as shown in Fig.~\ref{quant},~\ref{zoom}.} The network for coarse motion estimation (Coarse-Net) is built upon the U-Net architecture~\cite{ronneberger2015u} followed by fully connected (FC) layers to estimate the transformation parameters. 
The Coarse-Net takes an optical flow $\mathcal{F}_{j\rightarrow i}$ from image pairs $(I_j, I_i)$ and produces the transformation parameters that can relieve abrupt motion.
The network is shown in Fig.~\ref{overview}~(a). 

With this procedure, an $i$-th image $I_i$ has the estimated transformation parameters $\mathbf{p}_{j\rightarrow i}$ of the adjacent time stamps $j\in\mathcal{N}$. They are averaged to produce a transformation matrix $\textrm{T}$.
It is obtained from $\overline{\mathbf{p}}_i$ as defined below:
\begin{equation}
    \overline{\mathbf{p}}_i:=\frac{1}{2N}\sum_{j = -N, j\neq i}^{N}\mathbf{p}_{i+j\rightarrow i}.
    \label{avg}
\end{equation}
$\textrm{T}$ adjusts the target frame to produce $C_i=\textrm{T}(I_i)$ for stabilization.\footnote{For convenience, we denote $\textrm{T}$ as a transformation matrix as well as a function that transforms image using $\textrm{T}$.} 
This procedure is equivalent to applying the moving average filter~\cite{smith1997scientist} to the relative displacements between frames within the window $2N$. 
Since the moving average filter is a low-pass filter, it leads to noise suppression effects.
This procedure enables finding the transformation to stabilize the target frame, while avoiding expensive camera pose estimation and optimization.

\vspace{-2mm}
\subsection{Fine stabilizer}
\label{finestab}
\vspace{-2mm}
The image transformations estimated by the coarse stabilizer alone is not enough to handle parallax and spatially varying instability. To stabilize such spatially varying instability, we employ \emph{spatially smoothed optical flow}.
This stage is inspired by Liu \etal~\cite{liu2014steadyflow}'s work. It demonstrates spatially smoothed pixel trajectories are what essentially underlies between stabilized frames. 

In practice, the smooth flow should be robust to foreground object movement. 
For example, a moving person in a scene cannot interfere with the optical flow that is induced solely by camera movement.
Prior methods implement the smooth optical flow by identifying pixels with discontinuous flow vectors~\cite{liu2014steadyflow}, which can be a slow process. Instead, we utilize a simple neural network generating the smoothed optical flow through a single feed-forward.

\begin{SCfigure}[][t]
    \centering
    \includegraphics[width=0.45\linewidth,keepaspectratio]{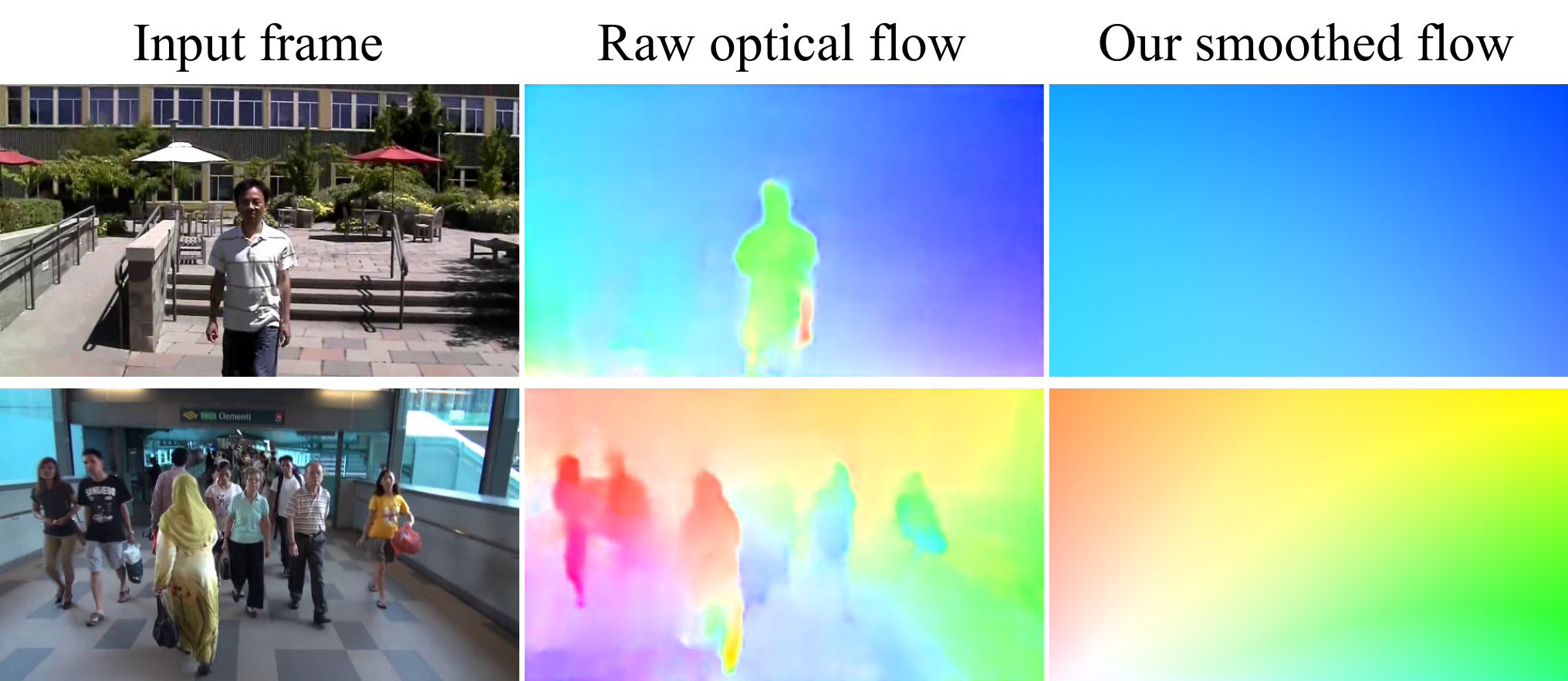}
    \caption{The Fine-Net produces spatially smoothed flow maps via a single feed forward operation that is purely induced by camera movement.
    The raw optical flow responds to the foreground, while our network is agnostic to it.
    }
    \label{flow}
    \vspace{-3mm}
\end{SCfigure}

In this regard, the architecture of the fine stabilizer (Fine-Net) produces smooth flow with a tiny U-shaped network. 
Fine-Net applies subsequent down-sampling of the features for filtering noisy motions. 
In addition, the average pooling layers in the network further induces spatial smoothness. 
Thus, by design, the Fine-Net does not produce high-frequency flow.
Fine-Net is shown in Fig.~\ref{overview}~(b), and examples of smooth flow are shown in Fig.~\ref{flow}.
It shows that output flow is agnostic to small foreground movement.\footnote{In Fig.~\ref{flow}, we use the full-sized images for the raw optical flow computation for the illustration. Fine-Net uses resized video instead for optical flow computation as discussed in~Sec.~\ref{sec:efficient_implementation}.}

Likewise in the coarse stabilizer, the moving average is applied to a group of smooth flow maps in a sliding window manner as Eq.~(\ref{avg}). 
Averaged flow map generated from a window of $C_i$ frames, produce a \emph{stabilizing warp} $\textrm{W}$ for frame $C_i$, resulting in $F_i$. 
Unlike the recent work by Yu and Ramamoorthi~\cite{yu2020learning}, our approach does not require an additional PCA component, and solves the problem with simple network architectures.

\vspace{-2mm}
\subsection{Margin inpainter}
\vspace{-2mm}
As frames are adjusted to stabilized positions, missing blank regions are inevitably created at the frame margins.
This is due to temporally unseen content during camera shakes.
Thus, a typical post-processing, cropping the boundary of frame was needed in previous approaches~\cite{liu2016meshflow,liu2013bundled,wang2018deep} to conceal missing blank regions.
However, excessive cropping leads to loss of original content and the zoom-in effect as a consequence.
We attempt to minimize such cropping via the margin inpainter.

Our approach takes five frames, namely $I_{i-2}$, $I_{i-1}$, $F_i$, $I_{i+1}$, and $I_{i+2}$, where $F_i$ is the warped image obtained from Fine-Net.
For motion compensation of the adjacent frames, the smooth flow maps estimated from the Fine stabilizer are used to warp adjacent frames to the stabilized center frame $F_i$. 
The aligned image frames are fed as input to the Margin-Net, as shown in Fig.~\ref{overview}~(c).

Inspired by the gated convolutions~\cite{yu2018free}, we implement the pixel gates via convolutional layers that act as spatial attention modules.
Adopting gated convolutions leads to effectively handling the arbitrary shapes of the boundary masks, which is beneficial compared to the equivariant nature of vanilla convolutional layers.

As a result, the Margin-Net learns how to combine adjacent frames to fill in the missing area of $F_i$, and produces the inpainted image $I_i^m$.
The Margin-Net concatenates the original $I_i^m$ and warped frames $F_i$ as input to generate the inpainted frame $O_i$.

\begin{figure*}
    \centering
    \includegraphics[width=0.995\linewidth,keepaspectratio]{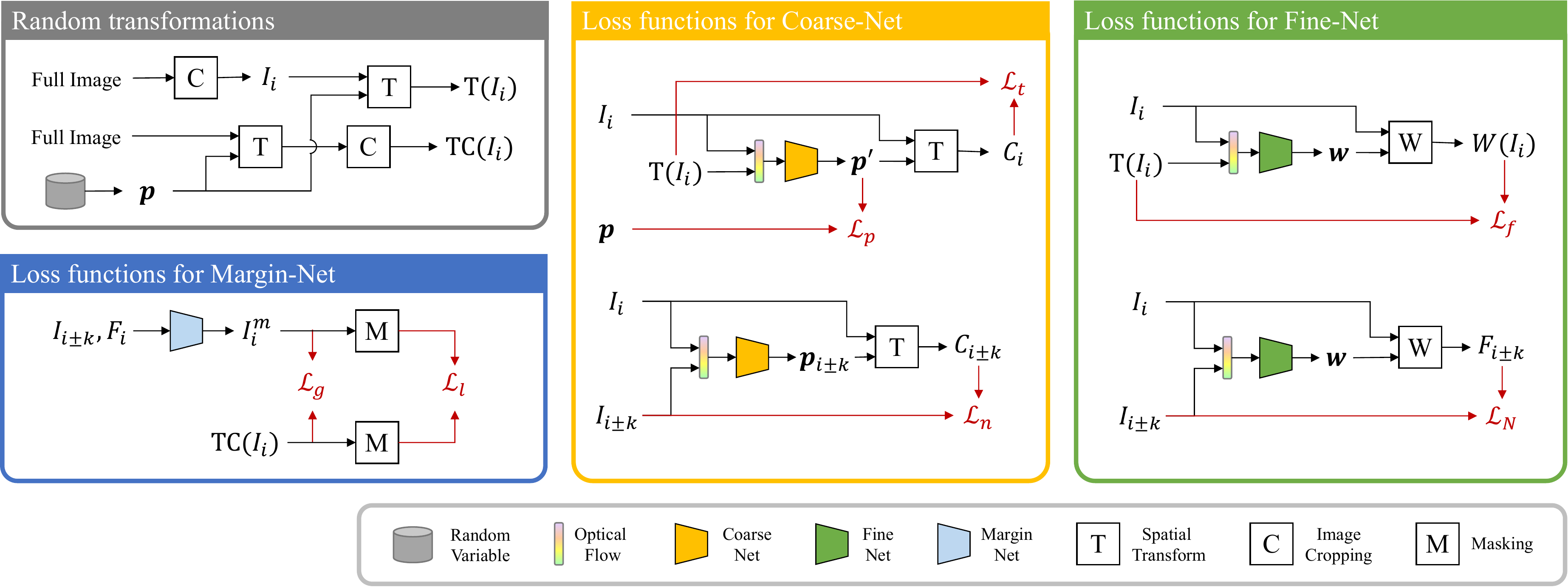}
    \caption{Overview of all loss functions for self-supervised training. 
    Losses for Coarse-Net (yellow box), Fine-Net (green box), and Margin-Net (blue box) are shown. 
    For each network, the original image $I_i$ is being self-augmented using the random image transforms (gray box), and it is used for the loss function shown in red. 
    For the data augmentation, 256$\times$256 patches are used for training from the original image in the training phase.
    }
    \vspace{-3mm}
    \label{loss}
\end{figure*}

\vspace{-3mm}
\section{Self-Supervised Learning}
\vspace{-2mm}
To train Coarse-Net, Fine-Net, and Margin-Net appropriately, we propose a novel self-supervised learning scheme. This distinguishes our approach with supervised methods~\cite{wang2018deep,xu2018deep}.
We carefully design the training methods for each module\footnote{It is worth to note that all three networks in our pipeline can handle an arbitrary size of images -- Coarse-Net and Fine-Net take rescaled flow fields as input, and Margin-Net is fully convolutional. 
With this in mind, we use smaller patches of $256 \times 256$ pix. of the video frames, to augment data and to reduce memory consumption during training. 
The cropped frames can be flipped horizontally or vertically for further augmentations.}, which is a crucial aspect of making our approach work. An overview of the loss functions is depicted in Fig.~\ref{loss}.

\vspace{-3mm}
\subsection{Training Coarse-Net}
\label{sec:training_coarse}
To train the Coarse-Net, we generate a pair of an original image and a randomly transformed image. 
The random transformation parameters $\mathbf{p}$ (e.g., rotation, translation) is sampled from a bounded uniform distribution ranging up to 30 degrees and 50 pixels respectively.

\vspace{2mm}
\noindent\textbf{Learning by random augmentation.}
As the first step, the image $I_i$ is transformed via the random rigid transformation matrix, yielding $\mathrm{T}(I_i)$.
Then, the flow map between $I_i$ and $\mathrm{T}(I_i)$ is computed and fed to the Coarse-Net which estimates the parameter vector $\mathbf{p}'$.
$I_i$ is then transformed with $\mathbf{p}'$ and results in $C_i$.
We use the L1 loss function between the randomly transformed frame $\mathrm{T}(I_i)$ and the estimated transformed frame $C_i$ defined as: 
\begin{equation}
{\color{red}\mathcal{L}_t}=\left\| \mathrm{T}(I_i) - C_i \right\|_1    
\label{loss1c}
\end{equation}

In addition, since we know the exact (randomly generated) ground truth (GT) parameter values, we can penalize the discrepancy between the GT parameter vector $\mathbf{p}$ and the estimated vector $\mathbf{p}'$. The loss is ${\color{red}\mathcal{L}_p}=\left\| \mathbf{p} - \mathbf{p}' \right\|_1$.

\vspace{2mm}
\noindent\textbf{Learning by predicting rigid transformation.}
$\mathcal{L}_t$ and $\mathcal{L}_p$ show losses using a rigidly transformed version of the same frame. In addition to these synthetic augmentations, another type of loss teaches how to roughly align temporally adjacent image pairs.

For this purpose, we collect adjacent frames of $I_i$ towards each of the $K$ neighboring frames from both sides\footnote{We use $K=2$ in our experiment.}, such as $I_{i-K},...,I_{i-1}$ and $I_{i+1},...,I_{i+K}$.
In this training phase, the optical flow is computed via inputs $(I_i, I_{i+k})$, and the Coarse-Net estimates the transform parameters $\mathbf{p}_{i+k}$. $\mathbf{p}_{i+k}$ transforms $I_i$ and produces $C_{i+k}$.
Now we apply the sum of L1 losses using $I_i$ and $C_{i+k}$,
${\color{red}\mathcal{L}_n}=\sum\limits_{k\in\pm[1,K]}\left\| I_{i+k} - C_{i+k} \right\|_1$.
This loss function makes Coarse-Net to learn rigid transform parameters for roughly aligning adjacent frames. 
This loss term increases the robustness for handling dynamic scenes.

Thus, the overall training loss for the Coarse-Net is the weighted sum of the three losses $L_C = \alpha \mathcal{L}_t + \beta \mathcal{L}_p + \gamma \mathcal{L}_n$, where $\alpha = \beta = 1.0$ and $\gamma = 0.1$.

\vspace{-3mm}
\subsection{Training Fine-Net}
\label{sec:training_fine}

\noindent\textbf{Learning by random augmentation.}
Training Fine-Net is similar to the training steps for Coarse-Net. We reuse the randomly transformed image $\mathrm{T}(I_i)$ in Eq.~\eqref{loss1c} and compute the optical flow between $\mathrm{T}(I_i)$ and $I_i$.
This flow map is fed through the Fine-Net, producing a \emph{smoothed flow map}.
Frame $I_i$ is then warped by performing the backward warping technique, using the computed smooth flow, producing the warped frame $\textrm{W}(I_i)$.
The L1 loss is applied between $\mathrm{T}(I_i)$ and the estimated frame $\textrm{W}(I_i)$, such as ${\color{red}\mathcal{L}_f}=\left\| \mathrm{T}(I_i) - \textrm{W}(I_i) \right\|_1$.

This loss helps the network to learn a relatively simple warping map which provides a warm start for training.
Note that the \emph{spatially smoothed} flow maps are produced by the Fine-Net's architecture design, as shown in Fig.~\ref{overview} (b).

\vspace{2mm}
\noindent\textbf{Learning by predicting smoothed flow.}
The Fine-Net aims to estimate smooth flow maps between neighboring frames and $I_i$.
We apply the sum of L1 losses to the $K$ frames as follows:
${\color{red}\mathcal{L}_N}=\sum\limits_{k\in\pm[1,K]}\left\| I_{i+k} - F_{i+k} \right\|_1$,
where $F_{i+k}$ is the frame $I_i$ warped towards $I_{i+k}$ using the Fine-Net's predicted smoothed flow.
The overall training loss for the Fine-Net is the weighted sum of the two losses $\mathcal{L}_F = \alpha \mathcal{L}_f + \gamma \mathcal{L}_N$, where $\alpha = 1.0$ and $\gamma = 0.1$.

\vspace{-3mm}
\subsection{Training Margin-Net}
\label{sec:training_margin}

The task of the Margin-Net is to fill in natural content in the blank margins of $F_i$, as shown in Fig.~\ref{overview} (c). 
To train this network, we prepare a pair of images, with or without blank margins.
In order to obtain such images, we reuse $\mathrm{T}(I_i)$ in Eq.~(\ref{loss1c}).
In addition, the same random transform is applied to the \emph{full-sized} image of the dataset, and then cropping is applied. 
In contrast to $\mathrm{T}(I_i)$, this procedure \emph{preserves} the content at the frame boundaries, as shown in Fig~\ref{loss}.
Let us call this image $\textrm{TC}(I_i)$.

\vspace{2mm}
\noindent\textbf{Learning to inpaint.}
Using the stabilized frame $F_i$ and warped images of $[I_{i-2}$, $I_{i-1}$, $I_{i+1}$, $I_{i+2}]$ as input, the Margin-Net outputs an inpainted frame $I_i^m$, utilizing the information from the given adjacent frames.
The Margin-Net is trained by the L1 loss between the transformed frame $\textrm{TC}(I_i)$ and the inpainted image $I_i^m$:
${\color{red}\mathcal{L}_g}=\left\| \textrm{TC}(I_i) - I_i^m \right\|_1$.

Furthermore, we can isolate the loss to focus only on the inpainted region by applying a mask to both $\textrm{TC}(I_i)$ and $I_i^m$:
${\color{red}\mathcal{L}_l}=\left\| m \odot (\textrm{TC}(I_i) - I_i^m) \right\|_1$,
where $m$ is the inpainting margin mask and $\odot$ denotes the element-wise product.
Since the stabilized frame $F_i$ is used as input, a mask indicating the inpainting region can be computed via the estimated transform parameter and flow map from the preceding Coarse and Fine-Net.
Using these two losses together increases the training stability: $\mathcal{L}_M=\mathcal{L}_g+\mathcal{L}_l$.

\vspace{-5mm}
\section{Experiments}
For thorough evaluation, we conduct an extensive quantitative comparison to both offline and state-of-the-art real-time methods, visual comparisons, analysis against a commercial product that runs offline, and ablation studies.
We also present results on full resolution videos, namely nHD (640$\times$360), HD (1280$\times$720), and FHD (1920$\times$1080).
For approaches without public code, we refer to the reported numbers in each paper.
For quantitative comparison, we use three commonly used metrics~\cite{liu2017codingflow,liu2016meshflow,liu2013bundled,wang2018deep,xu2018deep} to assess video stabilization quality, namely the \emph{cropping ratio}, \emph{distortion value}, and \emph{stability score}.
For details including implementation details, please refer to the supplementary material.

\vspace{1mm}\noindent
\noindent\textbf{Test videos.} Prior arts use videos that are publicly available from Liu~\etal~\cite{liu2013bundled}. 
However, it is important to note that \emph{no prior work validates all videos} provided by Liu~\etal~\cite{liu2013bundled} for video stabilization. 
Instead, each method uses an arbitrary subset of the videos. 
To configure the same experimental setup and to respect the numbers reported by state-of-the-arts, we use the \emph{union} of the video sets that were tested by them~\cite{goldstein2012video,grundmann2011auto,liu2009content,liu2011subspace,liu2017codingflow,liu2016meshflow,liu2013bundled,liu2014steadyflow,wang2018deep,xu2018deep}. Therefore, some videos that are not tested by an approach is not displayed in this paper. 

\begin{figure}[t]
    \centering
    \includegraphics[width=0.9\linewidth]{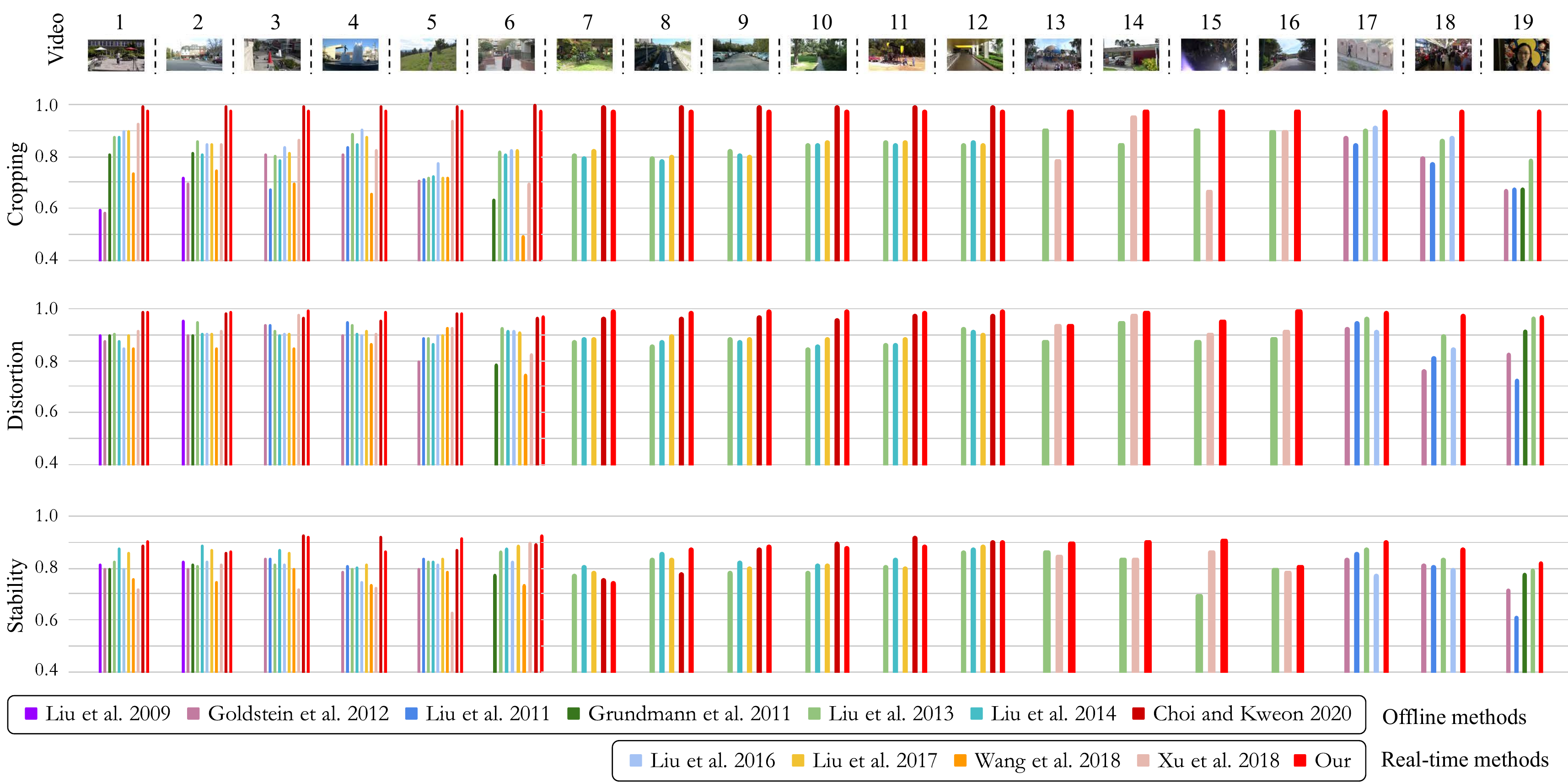}
    \caption{Quantitative comparison with the state-of-the-art video stabilization methods~\cite{goldstein2012video,grundmann2011auto,liu2009content,liu2011subspace,liu2017codingflow,liu2016meshflow,liu2013bundled,liu2014steadyflow,wang2018deep,xu2018deep}. 
    The results are evaluated with the three metrics: cropping ratio, distortion value, and stability score. 
    Higher values indicate better performance.
    }
    \label{quant}
    \vspace{-2mm}
\end{figure}

\begin{figure}[t]
    \centering
    \includegraphics[width=0.8\linewidth]{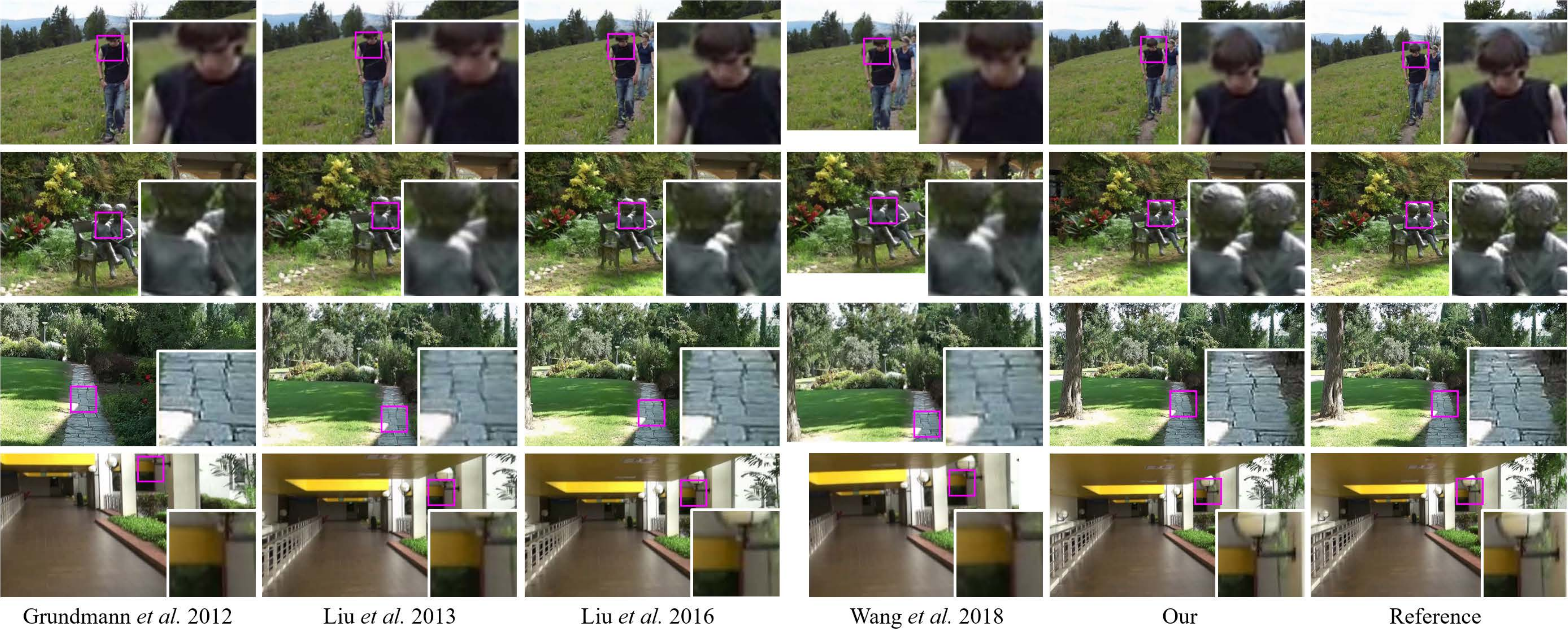}
    \caption{Visual comparison with \emph{offline} approaches~\cite{grundmann2012calibration,liu2016meshflow,liu2013bundled,wang2018deep} in terms of zoom-in effect. 
    Pink bounding boxes of the same size are magnified to show image regions at the same respective locations.
    Our approach shows the least amount of distortion and zoom-in effect.
    Please see our supplementary video for visual results.
    }
    \label{zoom}
    \vspace{-3mm}
\end{figure}

The union of video set consists of 19 video clips, and it covers forward/backward motion (\#1-5, 9, 10, 18), side motion (\#6, 7, 11, 12, 14) zoom-in and zoom-out (\#8, 17), the rolling-shutter artifact (\#13, 15, 16, 19).
The thumbnails of the video clips are shown in Fig.~\ref{quant}.
Note that our approach is validated with the union of the video sets tested by all prior works, and we emphasize that this is the \emph{most extensive collection of comparisons} in the video stabilization literature.
Furthermore, our network is trained on the DAVIS dataset~\cite{Perazzi2016} different from the 19 test clips, which verifies the generalization of our approach.

\vspace{1mm}\noindent
\textbf{Evaluation result.}
As shown in Fig.~\ref{quant}, the comparison with offline methods~\cite{goldstein2012video,grundmann2011auto,liu2009content,liu2011subspace,liu2013bundled,liu2014steadyflow}, online (real-time) methods~\cite{liu2017codingflow,liu2016meshflow,wang2018deep,xu2018deep}, and ours indicates that the proposed method shows favorable performance for the majority of videos. In particular, our real-time method shows comparable cropping ratio to the recent approach proposed by Choi and Kweon~\cite{choi2020deep}, although their approach runs offline and is designed to prevent any cropping. 

We also conduct comparisons with a widely used commercial product, Adobe Premiere (Pro CC 2017), and two recent approaches by Choi and Kweon~\cite{choi2020deep} and by Yu and Ramanmoorthi~\cite{yu2020learning}.
Please refer to the supplementary material.

\vspace{1mm}\noindent
\textbf{Ablation study.}
We conduct an extensive analysis of stabilization scores with various combination of proposed modules (Coarse-Net, Fine-Net, Margin-Net), window sizes, regarding resizing the frame inputs, types of transformations for Coarse-Net (translation, rotation, scale, and shear). Please see the supplement for the details.

\vspace{1mm}\noindent
\textbf{Visual comparison.}
Although it is not fair to directly compare our real-time method to offline methods, we present visual comparisons to the state-of-the-art offline methods~\cite{grundmann2012calibration,liu2016meshflow,liu2013bundled,wang2018deep} in Fig.~\ref{zoom}.
Notice that our method closely resembles the input content, while state-of-the-art methods convey enlargements due to the zoom-in effect.
Furthermore, we can observe moderate levels of blur introduced by the other methods, whereas our approach does not exhibit such artifacts.
For detailed visual results, please refer to the supplement.

\vspace{-5mm}
\section{Conclusion}
\vspace{-2mm}

We propose an unsupervised learning algorithm for video stabilization that runs in real-time.
Our approach consists of simple and efficient modules, such as Coarse-Net for image transform estimation, and Fine-Net for estimating the smoothed flow, and Margin-Net to compensate cropped contents. 
By combining all the modules, our approach outperforms both offline and real-time state-of-the-art methods on various videos (that shows zoom, rotation, parallax, and so on) and is robust to severe camera movements.


\vspace{-3mm}
\section{Acknowledgement}
This work was supported by the Institute for Information \& Communications Technology Promotion (2017-0-01772) grant funded by the Korea government.
This work was also supported by Institute of Information and communications Technology Planning and evaluation (IITP) grant funded by the Korea government (MSIT) (No.2021-0-02068, AI Innovation Hub)

\balance
{\small
\bibliography{egbib}
}

\end{document}